\documentclass{article}

\PassOptionsToPackage{sort&compress}{natbib} %
\usepackage[preprint]{corl_2026} %

\usepackage{myconfig}
\usepackage{pifont}
\usepackage{amsmath}

\definecolor{diffcolor}{HTML}{A61E5A} %
\newif\ifshowdiff
\showdifffalse
\newcommand{\diff}[1]{{\ifshowdiff\color{diffcolor}\fi#1}}
\newcommand{\diffon}{\ifshowdiff\color{diffcolor}\fi}
\renewcommand{\mai}[1]{#1}
\renewcommand{\yuri}[1]{#1}
\renewcommand{\jeremy}[1]{#1}
\usepackage{url}
\usepackage{wrapfig}
\usepackage{fontawesome5}

\AtBeginDocument{\setlength{\abovedisplayskip}{5pt}\setlength{\belowdisplayskip}{5pt}}

\hypersetup{
  pdftitle={CLAP: Direct VLM-to-VLA Adaptation via Language-Action Grounding},
  pdfauthor={Yuri Ishitoya, Jeremy Siburian, Masashi Hamaya, Haruto Suzuki, Kuniaki Saito, Toshihiko Fukushima, Cristian C. Beltran-Hernandez, Mai Nishimura},
  pdfsubject={},
}

\title{CLAP: Direct VLM-to-VLA Adaptation via Language-Action Grounding}

\author{
  Yuri Ishitoya$^{1,2,\dagger,\ddagger}$, Jeremy Siburian$^{1,3,\dagger,\ddagger}$, Masashi Hamaya$^{1}$, Haruto Suzuki$^{1,\dagger}$ \\
  \textbf{Kuniaki Saito$^{1}$, Toshihiko Fukushima$^{1}$, Cristian C. Beltran-Hernandez$^{1}$, Mai Nishimura$^{1,\ddagger}$} \\[4pt]
  \diff{$^{1}$OMRON SINIC X Corporation \quad $^{2}$Ochanomizu University \quad $^{3}$The University of Tokyo} \\
  \texttt{mai.nishimura@sinicx.com} \\[2pt]
  {\small $^{\dagger}$Work done during an internship at OMRON SINIC X. \quad $^{\ddagger}$Equal contribution.}
}

\begin{document}
\maketitle

\begin{abstract}
Vision-language-action models (VLAs) inherit semantic capabilities from pretrained VLMs, yet large-scale post-training on robot data and architectural modifications can reshape the backbone so extensively that it becomes difficult to isolate what the VLM contributes to control.
Directly converting pretrained VLMs into VLAs with minimal architectural change
offers a more transparent path to understanding how VLM capabilities transfer across model scales.
The core obstacle is output-distribution mismatch: predicting actions as bare numeric token sequences moves generation away from the VLM's pretrained language distribution, degrading the capabilities we seek to preserve.
To address this, we propose \textbf{CLAP} (\textbf{C}ausal \textbf{L}anguage-\textbf{A}ction \textbf{P}rediction), which prepends each numeric action sequence with a natural-language action description, causally conditioning precise action-token prediction on a language-action plan without modifying the backbone architecture.
\diff{With only 4--5 hours of training,} 2B CLAP achieves 90.8\% on LIBERO (+14.9\,pt over VLA-0) and improves robustness on LIBERO-PRO under language, object, and spatial perturbations.
\diff{Ablations attribute the gains to action-grounded language semantics, and on a real UR5e robot 2B CLAP reaches 60\% success against 15\% for matched VLA-0.}
We release CLAP at 0.8B, 2B, and 4B as an open-weight, multi-scale compact VLA family from a single VLM lineage, enabling controlled analysis of VLM-to-VLA capability transfer.
\par\noindent\diff{\faGlobe~Project page: \url{https://omron-sinicx.github.io/clap}}
\end{abstract}


\section{Introduction}

\mai{Rapid progress in vision-language models (VLMs) continually \jeremy{expands the capabilities of robot \textit{perception} and \textit{reasoning}}, yet transferring these capabilities to robot \textit{control} remains disproportionately difficult. Vision-language-action models (VLAs), which sit at the intersection of vision-language modeling and robot learning, pursue this transfer by fine-tuning pretrained VLMs on robot demonstrations~\cite{nvidia2025gr00tn1openfoundation,black2024pi0visionlanguageactionflowmodel,zha2026laplanguageactionpretrainingenables}. 
Recent systems have improved action prediction through \jeremy{discretized action tokens, learned action experts, or diffusion action heads}~\cite{nvidia2025gr00tn1openfoundation,kim2024openvlaopensourcevisionlanguageactionmodel}, but each addition \jeremy{moves the VLA further from its underlying VLM in both architecture and training distribution}. This architectural and data complexity makes it difficult to isolate what the pretrained VLM actually contributes to control, limiting both systematic capability analysis and the transfer of advances from the broader VLM community. \emph{What would it take to make VLA research as accessible as VLM research?}
}

\mai{A central source of complexity in \jeremy{VLM-to-VLA adaptation}}
is an \emph{output-distribution mismatch}: standard VLA fine-tuning trains a VLM pretrained to generate semantically structured language to instead emit bare action-token sequences such as \texttt{4 12 98 3 0 0}. These tokens carry little of the linguistic or spatial structure seen during pretraining, so learning to act can degrade the representations that support semantic generalization. This unresolved mismatch is especially important for \textit{compact} VLAs, which use lightweight VLM backbones for efficient deployment but may be more sensitive to distribution shift during fine-tuning~\citep{shukor2025smolvlavisionlanguageactionmodelaffordable,wen2025tinyvlafastdataefficientvisionlanguageaction}. Recent architecture-free VLM-to-VLA work provides cleaner settings for studying this issue by avoiding separate action experts, but existing approaches either predict bare action tokens~\citep{goyal2025vla0buildingstateoftheartvlas} or \diff{write the whole action hierarchy as text through a multi-stage pipeline}~\citep{hancock2025actionslanguagefinetuningvlms}. These choices leave a trade-off between \diff{direct, single-pass} executability and alignment with the VLM's pretrained language distribution.

\begin{figure}[t]
  \includegraphics[width=\linewidth]{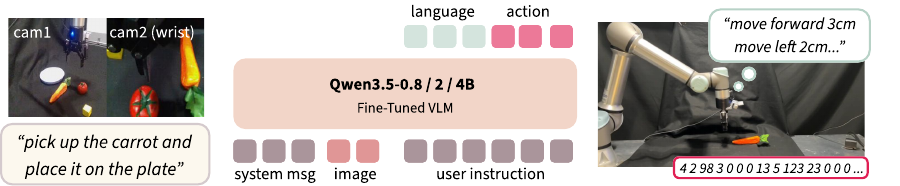}
   \caption{CLAP converts a pretrained VLM directly into a deployable VLA by prepending a natural-language action description to numeric action tokens, keeping every prediction step closer to the VLM pretraining distribution. CLAP requires no action expert or architectural change.}
   \label{fig:teaser}
\end{figure}

To address this mismatch, we propose \textbf{CLAP} (\textbf{C}ausal \textbf{L}anguage-\textbf{A}ction \textbf{P}rediction), a lightweight VLM-to-VLA fine-tuning recipe that \textbf{reformulates action prediction as language-conditioned action generation}. Instead of predicting bare action tokens directly or replacing actions with language-only descriptions, CLAP generates both in a single autoregressive sequence: before predicting discrete numeric action tokens, the model first generates a natural-language action description, such as \emph{``move forward, tilt right, and close the gripper''}. Because the full sequence is generated autoregressively, causal attention conditions the numeric action tokens on the preceding language description, making it a conditioning intermediate rather than an auxiliary prediction. CLAP requires \textbf{no action expert, vocabulary modification, architectural change, or manual annotation}, while retaining executable numeric action tokens for precise control.

We evaluate CLAP across manipulation, robustness, and capability-transfer analysis using compact Qwen3.5 backbones at 0.8B, 2B, and 4B parameters~\cite{qwen3.5}. On LIBERO, 2B CLAP achieves 90.8\% success with \diff{only 4--5 hours of training}, outperforming VLA-0 by +14.9 percentage points under identical conditions and approaching
\mai{fully-trained VLA-0's performance}, showing that the language-action prefix substantially improves learning efficiency. On LIBERO-PRO, CLAP improves robustness under instruction, object, and spatial perturbations, suggesting that the prefix helps beyond in-distribution imitation. On VLABench, the 2B backbone performs strongest across most capability dimensions and also yields the strongest CLAP policy, indicating that parameter count alone does not determine transfer quality in the compact regime. \diff{On a real UR5e robot, 2B CLAP also outperforms matched VLA-0.} Together, these evaluations suggest that CLAP provides not only an efficient compact VLA recipe, but also a controlled setting for studying how VLM capabilities transfer to robot control.

In summary, our main contribution is \textbf{CLAP}: a minimal VLM-to-VLA recipe that changes only the output representation, jointly generating language-action descriptions and action tokens without an action expert, vocabulary \jeremy{modification}, or architectural \jeremy{change}. 
\mai{Consistent with the lightweight design, CLAP reaches $\approx$ 90\%
success rate on LIBERO within 6.5 hours on 8 GPUs, across 0.8B, 2B, and 4B model sizes.}
We \diff{release} weights, code, and training recipes to support future research on VLM-to-VLA adaptation.

\section{Related Work}
\label{sec:related_work}

\paragraph{Vision-language-action models and action representations.}
Vision-language-action (VLA) models adapt pretrained vision-language models (VLMs)~\citep{karamcheti2024prismaticvlmsinvestigatingdesignspace,liu2023visualinstructiontuning,beyer2024paligemmaversatile3bvlm} to robot control by combining internet-scale vision-language pretraining with behavioral cloning on robot demonstrations~\citep{brohan2022rt1roboticstransformerrealworld,brohan2023rt2visionlanguageactionmodels,octomodelteam2024octoopensourcegeneralistrobot}. A central design choice is how to represent low-level actions within or alongside the VLM. One strategy discretizes continuous actions into tokens, casting robot control as autoregressive next-token prediction~\citep{kim2024openvlaopensourcevisionlanguageactionmodel,goyal2025vla0buildingstateoftheartvlas,pertsch2025fast}\diff{, or writes action values as decimal strings to narrow the action-language gap~\citep{grover2025enhancinggeneralizationvisionlanguageaction}}. Another adds dedicated diffusion or flow-matching action heads to generate continuous actions directly~\citep{nvidia2025gr00tn1openfoundation,black2024pi0visionlanguageactionflowmodel,intelligence2025pi05visionlanguageactionmodelopenworld}. CLAP builds on the autoregressive direction, but changes the output representation itself: instead of predicting bare discrete action tokens, it inserts a language-action prefix before the action tokens without adding an action expert.

\paragraph{Language grounding and intermediate reasoning.}
Prior work has used language, affordances, and intermediate reasoning traces to make robot policies more semantically structured. One line of work conditions manipulation policies or high-level plans on natural language, often grounding instructions through learned affordances or embodied multimodal representations~\citep{stepputtis2020languageconditionedimitationlearningrobot,ahn2022doasincannotsaygrounding,huang2022innermonologueembodiedreasoning,driess2023palmeembodiedmultimodallanguage}. \diff{A closer line of work uses language as an intermediate representation for low-level control: RT-H~\citep{belkhale2024rth} predicts language motions such as ``move arm forward'' before actions, and VLM2VLA~\citep{hancock2025actionslanguagefinetuningvlms} writes subtasks, motion plans, and numeric actions all as text. CLAP is conceptually close to both, but its prefix is a quantitative summary of the action chunk that serves to keep generation close to the VLM's pretrained language distribution, emitted together with the numeric action tokens in a single autoregressive sequence.}
Embodied reasoning methods such as ECoT~\citep{zawalski2025roboticcontrolembodiedchainofthought}, ECoT-Lite~\citep{chen2025trainingstrategiesefficientembodied}, and TraceVLA~\citep{zheng2024tracevlavisualtraceprompting} rely on auxiliary intermediate representations, including reasoning tokens and visual traces, before action generation, but often incur additional supervision, prompting, or inference-time overhead.
LAP~\citep{zha2026laplanguageactionpretrainingenables} uses language-action descriptions as a pretraining signal, but still relies on an action expert for final action generation. By contrast, CLAP places the language-action description directly in the final autoregressive policy output, where it conditions subsequent action tokens while preserving executable numeric control.

\paragraph{Capability preservation.}
Fine-tuning VLMs on robot action data can disrupt pretrained representations, degrading capabilities such as multimodal understanding, spatial reasoning, and generalization~\citep{driess2025knowledgeinsulatingvisionlanguageactionmodels,grover2025enhancinggeneralizationvisionlanguageaction}. One line of work addresses this by insulating or preserving the VLM during policy learning. Knowledge Insulation~\citep{driess2025knowledgeinsulatingvisionlanguageactionmodels} blocks gradients from the action expert, FLOWER~\citep{pmlr-v305-reuss25a} decouples action generation from the VLM via an intermediate-fusion Flow Transformer, and GenVLA~\citep{grover2025enhancinggeneralizationvisionlanguageaction} \diff{pairs frozen and trainable vision encoders} with vision-language co-training. \diff{ReVLA~\citep{dey2025revla} gradually reverts the fine-tuned vision encoders to their original pretrained weights to recover robustness to out-of-domain objects and distractors, and MAPS~\citep{huang2025maps} keeps visual layers close to their pretrained weights while linearly relaxing this constraint toward the language layers.} These methods improve preservation, but introduce additional modules, training objectives, gradient-blocking schemes, \diff{proximity regularizers,} or co-training data. CLAP instead asks whether capability-preserving VLM-to-VLA adaptation can be improved by changing only the output representation, keeping the architecture and training objective unchanged.

\paragraph{Compact VLAs.}
Capability preservation is especially important for compact VLAs, where efficient deployment must be balanced against limited model capacity. SmolVLA~\citep{shukor2025smolvlavisionlanguageactionmodelaffordable}, TinyVLA~\citep{wen2025tinyvlafastdataefficientvisionlanguageaction}, and VLA-Adapter~\citep{wang2025vlaadaptereffectiveparadigm} improve efficiency through lightweight backbones, adapter-based tuning, or specialized policy modules. However, they either focus on a specific model scale or rely on additional policy/action components.
In contrast, CLAP applies one architecture-preserving recipe to 0.8B, 2B, and 4B Qwen3.5 backbones, enabling controlled analysis across scales.

\section{Preliminaries}
\label{sec:preliminaries}

Given an image observation $o$ and a natural-language instruction $l$, an autoregressive VLA models a robot policy $\pi_\theta$ by predicting an action sequence.
Each $h$-step action chunk $\bm{a} = (a_1, \dots, a_h)$ consists of $h$ sequential $7$-DoF commands, where each $a_t \in \mathbb{R}^7$ encodes end-effector translation, rotation, and gripper state.
Following prior work~\citep{goyal2025vla0buildingstateoftheartvlas}, each action dimension is discretized into $K=1000$ integer bins and represented as text tokens using the VLM's existing vocabulary.
The policy predicts the flattened action-token sequence autoregressively:
\begin{equation}
\pi_\theta(\bm{a} \mid o, l)
=
\prod_{i=1}^{7h}
\pi_\theta(a^{(i)} \mid a^{(<i)}, o, l),
\label{eq:vla0-policy}
\end{equation}
where $a^{(i)}$ denotes the $i$-th token in the flattened action sequence and $a^{(<i)}$ denotes all preceding action tokens.
The VLM backbone is thus fine-tuned with the same next-token prediction objective as language pretraining, while its outputs can be directly decoded into robot commands. However, the fine-tuning targets are bare integer sequences that lack the semantic structure of natural language, diverging from the distribution the backbone was pretrained to generate.
We refer to this shift as \emph{output-distribution mismatch}~\citep{zha2026laplanguageactionpretrainingenables,hancock2025actionslanguagefinetuningvlms}.
CLAP mitigates this mismatch by incorporating a language-grounded action description into the autoregressive output.

\section{CLAP: Causal Language-Action Prediction}
\label{sec:method}

\begin{figure}[t]
    \includegraphics[width=\linewidth]{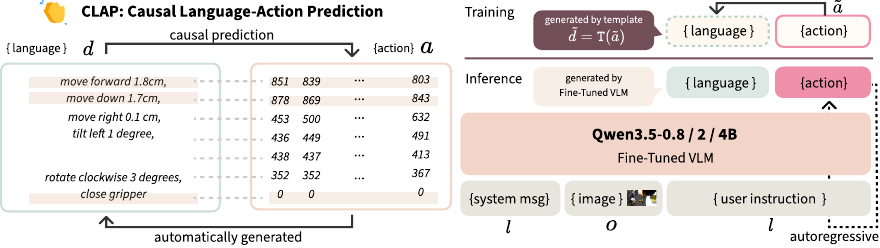}
  \caption{
  \textbf{Overview of CLAP.}
CLAP changes only the target output sequence for VLA fine-tuning. During training, the language-action description $\tilde{d}$ is generated from the ground-truth action chunk by a fixed template $\tilde{d} = \texttt{T}(a)$. At inference time, both language-action description $d$ and the numeric action tokens $a$ are generated autoregressively by the model. Because generation is causal, each action token conditions on the preceding language-action description.
  }
  \label{fig:clap}
\end{figure}

We use the output-distribution mismatch introduced in Section~\ref{sec:preliminaries} to design a simple training recipe for compact VLA policies. Our goal is to retain the semantic generalization of the pretrained VLM while producing directly executable low-level robot actions, without requiring an action expert, vocabulary extension, or large-scale robot pretraining. \Cref{fig:clap} provides an overview.

\paragraph{Language-action representation.}
The central design question in converting a pretrained VLM into a VLA is what \emph{action representation} to use. Representing actions as bare numeric tokens yields compact, directly executable outputs, but moves generation away from the VLM's pretrained language distribution. 
A promising alternative, introduced by LAP~\cite{zha2026laplanguageactionpretrainingenables}, is to express actions as natural-language descriptions, preserving closer alignment with the VLM's pretrained generation behavior. However, language-only action descriptions are too verbose to specify precise low-level commands efficiently, and still require an external action generator to produce executable robot actions~\cite{zha2026laplanguageactionpretrainingenables,driess2025knowledgeinsulatingvisionlanguageactionmodels}. 
\jeremy{CLAP addresses this trade-off by prepending a short language-action description to the numeric action tokens within a single autoregressive sequence, extending Eq.~\ref{eq:vla0-policy} as}
\jeremy{
\begin{equation}
\pi_\theta(d, \bm{a} \mid o, l)
=
\underbrace{
\prod_{j=1}^{|d|}
\pi_\theta\bigl(d^{(j)} \mid d^{(<j)}, o, l\bigr)
}_{\text{language-action prefix}}
\cdot
\underbrace{
\prod_{i=1}^{7h}
\pi_\theta\bigl(a^{(i)} \mid a^{(<i)}, d, o, l\bigr)
}_{\text{numeric action tokens}},
\label{eq:clap-policy}
\end{equation}
}
\mai{where $d$ is a natural-language description of the action chunk $\bm a$. During training, the description is generated deterministically from the ground-truth action chunk $\tilde {\bm a}$ by a fixed template as $\tilde d = \texttt {T}(\bm \tilde{a} )$. At inference time, the model generates both $d$ and $\bm a$ autoregressively.}
This representation satisfies three desiderata: (i) semantic alignment with the VLM's pretrained distribution, (ii) direct executability without an external action generator, and (iii) token efficiency for real-time control.

\paragraph{Template-based language-action prefix.}
We instantiate the language component of this representation with a deterministic template. For each $h$-step action chunk, CLAP prepends a compact natural-language summary of the intended motion, such as \texttt{"move forward 3 cm, tilt right 1 degree, close gripper"}. This prefix summarizes the chunk-level translation, rotation, and gripper command, while the following numeric tokens encode precise low-level action values. Because the prefix is generated from the ground-truth action chunk using a fixed template, CLAP requires no manual annotation beyond the original robot demonstrations. \Cref{tab:action-example} \diff{(Appendix~\ref{app:template})} shows an example output sequence generated from an input image observation and language instruction. \diff{The \texttt{<think>}\ldots\texttt{</think>} delimiters are not an architectural requirement: any deterministic delimiter would separate the prefix from the numeric tokens. We adopt them because they give an unambiguous parsing boundary and are built-in special tokens in recent VLMs such as Qwen3.5, avoiding BPE fragmentation.}

\paragraph{Joint generation and training.}
Given the template-derived prefix, CLAP trains the model to predict a single concatenated sequence: language-action description tokens followed by numeric action tokens. %
We optimize the standard autoregressive cross-entropy objective
\begin{equation}
  \mathcal{L}_{\text{CLAP}}
  =
  -\mathbb{E}_{(o,l,a)\sim\mathcal{D}}
  \left[
    \log \pi_\theta(\tilde d \mid o,l)
    +
    \log \pi_\theta(\tilde a \mid o,l,\tilde d)
  \right],
\end{equation}
where $\mathcal{D}$ is the robot demonstration dataset and $T$ is the fixed template. Because the sequence is generated autoregressively, each action token causally attends to the preceding language-action prefix, making the prefix a conditioning intermediate rather than an independent auxiliary output. CLAP uses the same VLM backbone, tokenizer, output head, and cross-entropy loss for both language and action tokens, introducing no action expert, auxiliary head, or vocabulary extension.

\paragraph{Action masking augmentation.}
We consider an optional augmentation that changes the teacher-forced input context without changing the CLAP target sequence. Specifically, we randomly replace a subset of input action tokens with a ``\texttt{?}" placeholder while keeping the original tokens as prediction targets. The model must therefore infer the masked values from the image observation, instruction, language-action prefix, and remaining action context. For each sample, the masking fraction is drawn from $\mathrm{Uniform}(0,p_{\max})$, with a 40\% probability of applying no masking. Full details are provided in Appendix~\ref{app:masking}.
We study the effect of masking in Section~\ref{sec:learning_efficiency}; unless otherwise stated, CLAP uses the unmasked configuration.

\section{Experiments}
\label{sec:experiments}

\subsection{Experimental Setup}
\label{sec:setup}

\paragraph{Benchmarks.}
We evaluate CLAP on three benchmarks targeting complementary aspects of VLA performance: (1) in-distribution manipulation, (2) out-of-distribution robustness, and (3) VLM capability analysis. LIBERO~\citep{liu2023libero} evaluates in-distribution manipulation under the standard simulated benchmark protocol; we report results on the four LIBERO suites, covering spatial reasoning, object manipulation, goal-conditioned behavior, and long-horizon tasks. LIBERO-PRO~\citep{zhou2025liberopro} evaluates out-of-distribution robustness by perturbing the original LIBERO tasks, including novel visual object instances, object relocation, instruction rephrasing, and task-configuration changes. VLABench~\citep{liu2024vlabench} evaluates language-conditioned robot reasoning capabilities; we use it to measure the base Qwen3.5 backbone's planning and spatial reasoning capabilities before robot fine-tuning, and to analyze how these capabilities relate to downstream VLA performance. \diff{We further validate CLAP on a real UR5e robot (Section~\ref{sec:real_robot}).} Full benchmark details are provided in Appendix~\ref{app:benchmarks}.

\paragraph{Comparisons.}
Our primary baseline is VLA-0, reproduced using the same Qwen3.5 backbones, LIBERO training data, and \diff{17k-gradient-step} training budget\footnote{\diff{The original VLA-0 code lacks a distributed sampler, so each of the 8 GPUs iterates over the full dataset and one of its ``epochs'' equals 8 passes over the data. We therefore report gradient steps: 17k steps correspond to 1 epoch in VLA-0's code and 8 epochs in our implementation.}}. This matched comparison isolates the effect of the language-action prefix. We also report results from large robot-pretrained VLAs, including OpenVLA~\citep{kim2024openvlaopensourcevisionlanguageactionmodel}, $\pi_0$~\citep{black2024pi0visionlanguageactionflowmodel}, and SmolVLA~\citep{shukor2025smolvlavisionlanguageactionmodelaffordable}, as contextual reference points. The {\color{refgray}gray} rows in Table~\ref{tab:libero} indicate methods evaluated under different backbones, training budgets, or robot-pretraining data, and therefore should not be interpreted as matched comparisons. \diff{To isolate the source of CLAP's gains, we additionally compare against the string-based action tokenization of Grover et al.~\citep{grover2025enhancinggeneralizationvisionlanguageaction}, which writes normalized action values as fixed-precision decimal strings without a prefix, and against two prefix controls trained with the CLAP recipe: a \emph{random} prefix of meaningless words length-matched to CLAP, and a \emph{numeric} prefix listing the per-axis displacements (\texttt{dx:\,+6.2, dy:\,-2.1, \ldots}) without linguistic form. We also instantiate CLAP on Qwen3-VL-2B~\citep{qwen3vl}.}

\paragraph{Implementation.}
We instantiate CLAP with Qwen3.5 backbones at 0.8B, 2B, and 4B parameters. All models are trained for \diff{17k gradient steps with a batch size of 16 per GPU on 8 GPUs (global batch 128)} \mai{on the LIBERO demonstration data across all four suites} with full-parameter fine-tuning, using the same action discretization and fine-tuning protocol as VLA-0. Unless otherwise stated, CLAP uses the language-action prefix without additional augmentation. We study random action masking as an optional augmentation in Section~\ref{sec:learning_efficiency}; the unmasked variant is used as the default configuration. 
Full hyperparameters and evaluation details are provided in Appendix~\ref{app:hyperparams}.

\subsection{Learning Efficiency Results}
\label{sec:learning_efficiency}

\begin{table}[t]
  \centering
  \scriptsize
  \caption{LIBERO benchmark ($h{=}8$).
    \hcb{Best} result per VLA-0 vs.\ CLAP pair under the same \diff{17k-step} protocol.
    {\color{refgray}Gray} rows are prior work under their original training protocols (contextual reference).}
  \label{tab:libero}
  \begin{tabular*}{\textwidth}{@{\extracolsep{\fill}}llll cccc l}
  \toprule
  \textbf{Model} & \textbf{Backbone} & \textbf{Size} & \textbf{VLA Pretrain}
    & \textbf{Spatial$\uparrow$}
    & \textbf{Object$\uparrow$}
    & \textbf{Goal$\uparrow$}
    & \textbf{Long$\uparrow$}
    & \textbf{Avg$\uparrow$} \\
  \midrule
  \rowcolor{magenta!20}\multicolumn{9}{l}{\textit{Ours (\diff{17k steps}, identical protocol)}} \\
  \quad VLA-0 (repro.)   & Qwen3.5 & 0.8B & --
    & 77.6 & 87.8 & 78.4 & 60.6 & 76.1 \\
  \quad CLAP (Ours)      & Qwen3.5 & 0.8B & --
    & \hcb{\textbf{92.6}} & \hcb{\textbf{97.8}} & \hcb{\textbf{88.4}}
    & \hcb{\textbf{79.6}} & \hcb{\textbf{89.6}} {\color{green}\textbf{(+13.5)}} \\
  \midrule
  \quad VLA-0 (repro.)   & Qwen3.5 & 2B & --
    & 77.8 & 86.6 & 77.0 & 62.4 & 75.9 \\
  \quad CLAP (Ours)      & Qwen3.5 & 2B & --
    & \hcb{\textbf{93.0}} & \hcb{\textbf{97.4}} & \hcb{\textbf{90.8}}
    & \hcb{\textbf{82.0}} & \hcb{\textbf{90.8}} {\color{green}\textbf{(+14.9)}} \\
  \midrule
  \quad VLA-0 (repro.)   & Qwen3.5 & 4B & --
    & 37.6 & 84.8 & 85.8 & 48.6 & 64.2 \\
  \quad CLAP (Ours)      & Qwen3.5 & 4B & --
    & \hcb{\textbf{88.0}} & \hcb{\textbf{97.4}} & \hcb{\textbf{86.6}}
    & \hcb{\textbf{67.6}} & \hcb{\textbf{84.9}} {\color{green}\textbf{(+20.7)}} \\
  \midrule\midrule
  \rowcolor{yellow!10}\multicolumn{9}{l}{\rv{\textit{Prior work (full training protocol)}}} \\
  \quad \rv{SmolVLA$^\dagger$}  & \rv{SmolVLM2}   & \rv{2.25B} & \rv{--}
    & \rv{93.0} & \rv{94.0} & \rv{91.0} & \rv{77.0} & \rv{88.8} \\
  \quad \rv{VLA-0$^\ddagger$}   & \rv{Qwen2.5-VL} & \rv{3B}    & \rv{--}
    & \rv{97.0} & \rv{97.8} & \rv{96.2} & \rv{87.6} & \rv{94.7} \\
  \quad \rv{$\pi_0.5$}          & \rv{PaliGemma}  & \rv{3B}    & \rv{\ding{51}}
    & \rv{98.8} & \rv{98.2} & \rv{98.0} & \rv{92.4} & \rv{96.85} \\
  \quad \rv{OpenVLA}             & \rv{Prismatic}  & \rv{7B}    & \rv{\ding{51}}
    & \rv{84.7} & \rv{88.4} & \rv{79.2} & \rv{53.7} & \rv{76.5} \\
  \bottomrule
  \end{tabular*}
  \vspace{2pt}
  \scriptsize
  $\dagger$ SmolVLA: community-scale robot pretraining; all sizes in App.~\ref{app:smolvla}.
  $\ddagger$ Original VLA-0~\citep{goyal2025vla0buildingstateoftheartvlas}: Qwen2.5-VL-3B, \diff{64 epochs (per its code)}.
\end{table}

\paragraph{Does the language-action prefix improve learning efficiency?}
Table~\ref{tab:libero} evaluates performance after \diff{17k training steps}, isolating the effect of the output representation under matched backbones, data, and training budget. CLAP consistently improves over VLA-0 across all model sizes, with average gains of +13.5, +14.9, and +20.7 points at 0.8B, 2B, and 4B, respectively. At 2B parameters, CLAP achieves the best overall result, reaching 90.8\% average success and outperforming VLA-0 by \textbf{+14.9 points}. The gains are broad across task suites: 2B CLAP improves Spatial from 77.8\% to 93.0\%, Object from 86.6\% to 97.4\%, Goal from 77.0\% to 90.8\%, and Long from 62.4\% to 82.0\%. For context, 2B CLAP also exceeds the reported averages of SmolVLA and OpenVLA and approaches stronger robot-pretrained systems such as $\pi_{0.5}$, although these comparisons are not controlled for backbone, data, training budget, or pretraining.

\paragraph{Does action masking provide additional gains?}
Table~\ref{tab:ablation} \diff{(Appendix~\ref{app:masking})} ablates whether action masking improves over the language-action prefix alone. Masking is not uniformly beneficial: it reduces average success by 3.9 points at 0.8B and 1.7 points at 2B, but improves the 4B model by 3.2 points, from 84.9\% to 88.1\%. The drop at smaller scales is mainly due to lower Long-horizon performance, suggesting that 0.8B and 2B models benefit from direct access to the full action-token context. In contrast, the 4B model appears better able to exploit the additional reconstruction constraint imposed by masking. These results suggest that action masking should be treated as a validation-dependent augmentation rather than a core component of CLAP.

\begin{table}[t]
    \diffon
    \centering
    \scriptsize
    \caption{\textbf{Prefix content and action tokenization ablation} on LIBERO (2B, \diff{17k steps}, $h{=}8$).
             Grover et al.~\citep{grover2025enhancinggeneralizationvisionlanguageaction}: string-based action tokenization without a prefix.
             \textit{Len.}: mean generated prefix length in Qwen3.5 tokens.
             We highlight the \hcb{best} result per column within the Qwen3.5 block.}
    \label{tab:prefix_ablation}
    \begin{tabular}{llllccccl}
    \toprule
    \textbf{Backbone} & \textbf{Method} & \textbf{Prefix} & \textbf{Len.}
      & \textbf{Spatial$\uparrow$}
      & \textbf{Object$\uparrow$}
      & \textbf{Goal$\uparrow$}
      & \textbf{Long$\uparrow$}
      & \textbf{Avg$\uparrow$} \\
    \midrule
    Qwen3.5 & VLA-0 (repro.)  & --            & 0  & 77.8 & 86.6 & 77.0 & 62.4 & 75.9 \\
    Qwen3.5 & Grover et al.   & --            & 0  & 78.4 & 91.8 & 83.6 & 70.8 & 81.1 \\
    \midrule
    Qwen3.5 & CLAP            & Random words  & 34 & 79.2 & 84.0 & 85.6 & 67.0 & 79.0 \\
    Qwen3.5 & CLAP            & Numeric       & 54 & 91.2 & \hcb{\textbf{98.2}} & \hcb{\textbf{90.8}} & 74.4 & 88.7 \\
    Qwen3.5 & CLAP (Ours)     & Language      & 37 & \hcb{\textbf{93.0}} & 97.4 & \hcb{\textbf{90.8}} & \hcb{\textbf{82.0}} & \hcb{\textbf{90.8}} \\
    \midrule
    Qwen3-VL & CLAP (Ours)    & Language      & -- & 87.0 & 94.8 & 91.2 & 77.8 & 87.7 \\
    \bottomrule
    \end{tabular}
\end{table}

\diff{\paragraph{Where do the gains come from?}
Table~\ref{tab:prefix_ablation} ablates action tokenization and prefix content at 2B. Improving the action tokenization alone has limited effect: the string-based representation of Grover et al. improves over VLA-0 (81.1 vs.\ 75.9) but remains far below CLAP (90.8). On the prefix side, a random-word prefix barely improves over VLA-0 (79.0), whereas a numeric prefix that retains the action content without linguistic form reaches 88.7 but trails CLAP where task complexity is highest (Long: 74.4 vs.\ 82.0). Prefix length does not explain this ordering: the numeric prefix is the longest, and the random prefix is length-matched to CLAP yet 11.8 points worse. The gains therefore stem from action-grounded semantics expressed in natural language, not from additional tokens or a longer output sequence. Finally, CLAP is architecture-agnostic: on Qwen3-VL-2B, which differs from Qwen3.5 in fusion architecture and tokenizer, it reaches 87.7\% average success.}

\subsection{Generalization and Capability Transfer Results}
\label{sec:generalization_capability}

\begin{table*}[t]
  \centering
  \scriptsize
  \setlength{\tabcolsep}{3pt}
  \caption{
  LIBERO-PRO OOD generalization across task suites and perturbation types.
  Results are suite-level average success rates.
  \textit{Obj}: novel visual instances; \textit{Pos}: object relocation;
  \textit{Sem}: instruction rephrasing; \textit{Task}: task-configuration change.
  We highlight the \hcb{best} and \hcs{second-best} results in each column.
  All models use Qwen3.5 and are trained for \diff{17k gradient steps} with $h{=}8$.
  }
  \label{tab:libero_pro_transposed}
  \resizebox{\textwidth}{!}{%
  \begin{tabular}{llcccc|cccc|cccc|cccc|c}
  \toprule
  & & \multicolumn{4}{c|}{\textbf{Spatial}}
    & \multicolumn{4}{c|}{\textbf{Object}}
    & \multicolumn{4}{c|}{\textbf{Goal}}
    & \multicolumn{4}{c|}{\textbf{Long}} & \\
  \textbf{Model} & \textbf{Size}
    & \textbf{Obj} & \textbf{Pos} & \textbf{Sem} & \textbf{Task}
    & \textbf{Obj} & \textbf{Pos} & \textbf{Sem} & \textbf{Task}
    & \textbf{Obj} & \textbf{Pos} & \textbf{Sem} & \textbf{Task}
    & \textbf{Obj} & \textbf{Pos} & \textbf{Sem} & \textbf{Task}
    & \textbf{Avg} \\
  \midrule
  \multirow{3}{*}{VLA-0 (repro.)}
    & 0.8B
    & 70.2 & \hcs{9.4} & 59.2 & 50.4
    & 84.4 & 0.0 & 85.8 & 4.6
    & 34.6 & 0.0 & 65.6 & 13.2
    & 27.8 & 0.0 & 38.0 & 9.0
    & 34.5 \\
    & 2B
    & 67.4 & 0.6 & 51.4 & 44.8
    & 80.6 & 0.0 & 74.8 & 3.4
    & 44.0 & \hcb{\textbf{1.8}} & 49.4 & 8.0
    & 31.8 & 0.0 & 60.6 & 8.6
    & 32.9 \\
    & 4B
    & 39.8 & 0.0 & 32.6 & 31.8
    & 81.0 & 0.0 & 86.8 & \hcb{\textbf{5.4}}
    & 52.6 & 0.0 & 59.2 & 12.0
    & 34.4 & 0.0 & 45.8 & 8.8
    & 30.6 \\
  \midrule
  \multirow{3}{*}{CLAP w/o masking (Ours)}
    & 0.8B
    & 90.0 & 2.4 & 68.0 & 53.8
    & 91.2 & 0.0 & 89.0 & \hcb{\textbf{5.4}}
    & 44.4 & 0.2 & 54.4 & 13.2
    & \hcb{\textbf{52.0}} & 0.0 & 64.4 & 11.2
    & 40.0 \\
    & 2B
    & 92.6 & 8.4 & 66.4 & \hcb{\textbf{65.8}}
    & \hcs{92.4} & 0.0 & \hcs{95.8} & \hcb{\textbf{5.4}}
    & \hcb{\textbf{57.6}} & \hcb{\textbf{1.8}} & \hcs{76.6} & 14.4
    & \hcs{48.2} & 0.0 & 68.2 & 10.2
    & \hcs{44.0} \\
    & 4B
    & 82.4 & 8.4 & \hcs{81.0} & 53.2
    & 91.4 & 0.0 & \hcb{\textbf{98.8}} & \hcb{\textbf{5.4}}
    & 38.2 & \hcs{1.0} & 68.0 & \hcb{\textbf{17.2}}
    & 39.8 & \hcb{\textbf{0.2}} & 64.2 & \hcb{\textbf{15.4}}
    & 41.5 \\
  \midrule
  \multirow{3}{*}{CLAP w/ masking (Ours)}
    & 0.8B
    & 75.4 & 0.0 & 48.4 & 52.2
    & \hcb{\textbf{93.8}} & 0.0 & 82.0 & \hcs{5.2}
    & 52.6 & 0.8 & 56.8 & \hcs{16.4}
    & 38.6 & 0.0 & 27.2 & 7.6
    & 34.8 \\
    & 2B
    & \hcs{93.8} & 6.6 & 75.4 & 52.2
    & 91.6 & 0.0 & 91.4 & 3.8
    & 46.0 & 0.6 & 71.8 & 10.4
    & 40.2 & \hcb{\textbf{0.2}} & \hcs{68.4} & 13.2
    & 41.6 \\
    & 4B
    & \hcb{\textbf{94.2}} & \hcb{\textbf{10.4}} & \hcb{\textbf{88.2}} & \hcs{64.8}
    & 87.2 & 0.0 & 91.2 & 4.8
    & \hcs{56.6} & 0.0 & \hcb{\textbf{79.4}} & 11.4
    & 42.0 & 0.0 & \hcb{\textbf{72.2}} & \hcs{15.0}
    & \hcb{\textbf{44.8}} \\
  \bottomrule
  \end{tabular}
  }%
\end{table*}

\paragraph{Do CLAP's learning gains transfer out of distribution?}
Table~\ref{tab:libero_pro_transposed} assesses CLAP's robustness to out-of-distribution perturbations on LIBERO-Pro. CLAP consistently improves over VLA-0 across model scales, with average OOD gains of \jeremy{+5.5, +11.1, and +10.9 points for unmasked CLAP at 0.8B, 2B, and 4B, respectively.} The most striking improvement appears on the Spatial suite with novel visual instances. 
For 4B models, unmasked CLAP improves over VLA-0 by \textbf{+42.6\,pt}, while the masked variant \jeremy{improves by} \textbf{+54.4\,pt}. These results \jeremy{suggest that CLAP improves OOD robustness beyond what is captured by standard in-distribution evaluation on LIBERO.} Notably, action masking substantially enhances OOD robustness at 4B scale, \jeremy{despite providing only mixed benefits on LIBERO.}
This suggests that masking \jeremy{can reduce reliance on action-token shortcuts that become brittle under distribution shift.}
\diff{Notably, VLA-0's average OOD success \emph{decreases} with backbone size (34.5, 32.9, 30.6), consistent with the distribution-mismatch hypothesis: larger VLMs have more entrenched pretrained distributions, so \diff{17k steps} of bare action-token prediction degrades OOD-relevant knowledge more at larger scale. CLAP restores positive scaling from 0.8B to 2B by reducing this shift, and masking adds regularization that benefits only the over-parameterized 4B regime.}

\begin{figure}[t]
  \centering
  \includegraphics[width=0.8\linewidth]{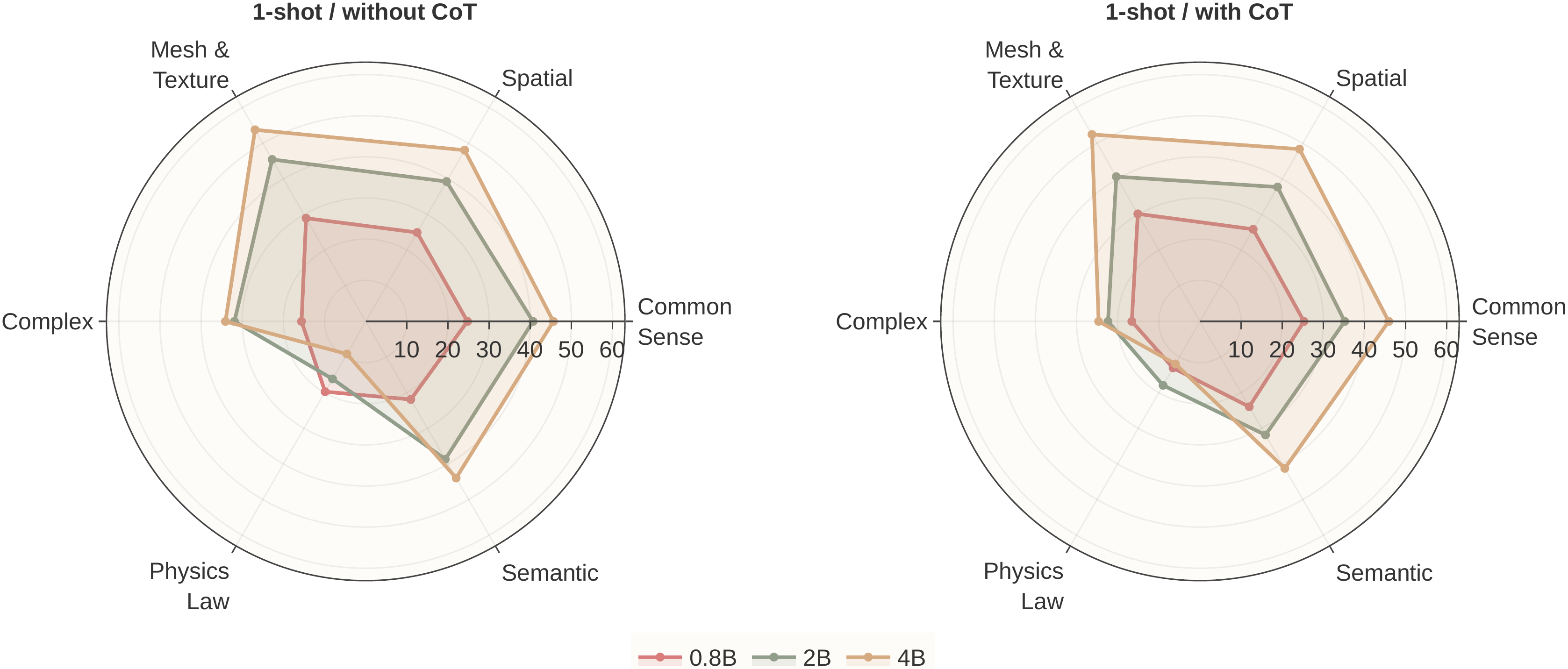}
  \caption{\textbf{VLABench profiles \textit{before} robot fine-tuning} (Qwen3.5 0.8B/2B/4B, 1-shot, without and with CoT). 2B is strong and balanced; CoT mainly helps 4B on Spatial and Mesh \& Texture.}
  \label{fig:vlabench_main}
\end{figure}
\paragraph{Does bigger VLM capability mean better transfer?}
Figure~\ref{fig:vlabench_main} presents VLABench performance of Qwen3.5 backbones prior to robot fine-tuning, evaluated under 1-shot prompting with and without \jeremy{Chain-of-Thought (CoT)}  \citep{wei2022chainofthoughtpromptingelicitsreasoning}. Across most categories, performance generally improves with model size. However, on Complex and Physics Law tasks, the 0.8B and 2B models match or exceed the 4B model under both prompting conditions. This \jeremy{non-monotonic trend} persists downstream: CLAP 2B also outperforms CLAP 4B on LIBERO (Table~\ref{tab:libero}). Base VLM capability thus influences VLA performance, yet parameter count alone does not determine transfer quality \jeremy{in the compact regime.}

\begin{wrapfigure}[15]{r}{0.44\linewidth}
  \vspace{-12pt}
  \centering
  \diff{\includegraphics[width=\linewidth]{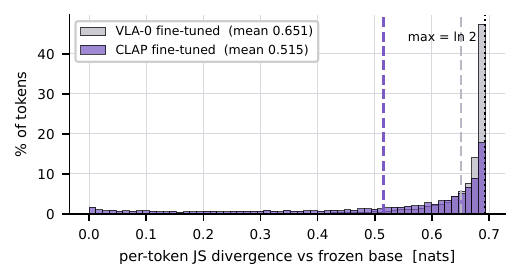}
  \caption{\textbf{Token-level distribution shift from the pretrained VLM.} Per-token JS divergence between frozen Qwen3.5-2B and each fine-tuned model on generic VQA prompts (256 held-out images, 17k-step checkpoints; $\ln 2$ is the maximum).}
  \label{fig:js_divergence}}
  \vspace{-10pt}
\end{wrapfigure}
\paragraph{Does reasoning help compact VLAs?}
Reading the same evaluation along the CoT axis, we find that removing CoT improves performance in most settings; on Physics Law tasks, the 0.8B model even exceeds the larger 2B and 4B models. \jeremy{Output analysis suggests that longer reasoning traces can introduce format violations or loops that diverge from actionable predictions, consistent with recent reports that extended reasoning may weaken visual grounding~\citep{raghu2026dontblinkevidencecollapse}.} CLAP \jeremy{avoids relying on free-form reasoning at inference time}: its structured language-action template \jeremy{provides} a compact intermediate representation \jeremy{before numeric action prediction}, keeping smaller variants competitive and the approach attractive for compact, deployment-friendly settings.

\paragraph{\diff{Does CLAP stay closer to the pretrained VLM?}}
\diff{On 256 held-out images paired with generic VQA prompts (never the action prompt), we greedy-decode a reference response from the frozen Qwen3.5-2B and compute the Jensen--Shannon (JS) divergence between the frozen model's and each fine-tuned model's next-token distributions along it (Figure~\ref{fig:js_divergence}). CLAP stays closer on all 256 samples (mean JS 0.515 vs.\ 0.651 for VLA-0; Wilcoxon $p<10^{-43}$), and only 32.9\% of its tokens are nearly disjoint from the base distribution (JS $>0.95\ln 2$) against 68.7\% for VLA-0. Enabling the chat template's thinking mode leaves the gap unchanged (0.525 vs.\ 0.658). Two limits apply: both models move far from the pretrained VLM in absolute terms, so CLAP degrades less rather than staying close, and restricted to the base model's top-50 tokens the two are indistinguishable, so CLAP's advantage lies in the distributional tail, where VLA-0 collapses onto digit tokens.}

\diff{\subsection{Real-Robot Evaluation and Inference Latency}
\label{sec:real_robot}
We deploy CLAP on a UR5e arm with a Robotiq 2F-85 gripper for pick-and-place in cluttered tabletop scenes (Appendix~\ref{app:real-robot}). The policy consumes a side-camera and a wrist-camera image and outputs an eight-step chunk of end-effector deltas and gripper state, executed open-loop. We collect 120 teleoperated demonstrations over three target objects with randomized layouts and instructions, and fine-tune Qwen3.5-0.8B and 2B from the pretrained VLM checkpoints without LIBERO pre-training. VLA-0 is trained with the same backbone, demonstrations, and protocol. Each model is evaluated over 20 trials in-domain (ID: seen objects and layouts) and out-of-domain (OOD: unseen distractor objects added). Table~\ref{tab:real_robot} shows that CLAP outperforms the matched VLA-0 baseline in both conditions (ID: 60.0\% vs.\ 15.0\%; OOD: 60.0\% vs.\ 5.0\% at 2B). VLA-0 is markedly less stable under minor variation in lighting and object placement, whereas 2B CLAP retains its in-domain success rate under distractors. The 0.8B model reaches 35.0\% ID but 10.0\% OOD, a wider capacity gap than on LIBERO. Absolute success rates remain below simulation with only 120 demonstrations, and we regard this study as preliminary.

\paragraph{Inference latency.}
Table~\ref{tab:latency} reports end-to-end latency per policy query on a single RTX~4090 ($h{=}8$). Under default eager decoding, the prefix adds about 0.4\,s per query at 2B, and the 0.8B and 2B models have near-identical latency, indicating that generation length rather than model size dominates at these scales. With \texttt{torch.compile}, CLAP-0.8B runs at 6.06\,Hz and CLAP-2B at 4.11\,Hz, which is practical for chunk-based control at $h{=}8$.}

\begin{table}[t]
  \diffon
  \scriptsize
  \begin{minipage}[t]{0.40\linewidth}
    \centering
    \caption{\textbf{Real-robot results} (\%; Qwen3.5, 20 trials per condition). ID: seen objects and layouts; OOD: unseen distractors added.}
    \label{tab:real_robot}
    \begin{tabular}{llcc}
    \toprule
    \textbf{Model} & \textbf{Size} & \textbf{ID$\uparrow$} & \textbf{OOD$\uparrow$} \\
    \midrule
    VLA-0 (repro.) & 2B   & 15.0 & 5.0 \\
    \midrule
    \multirow{2}{*}{CLAP (Ours)} & 0.8B & 35.0 & 10.0 \\
                   & 2B   & \hcb{\textbf{60.0}} & \hcb{\textbf{60.0}} \\
    \bottomrule
    \end{tabular}
  \end{minipage}\hfill
  \begin{minipage}[t]{0.56\linewidth}
    \centering
    \caption{\textbf{Inference latency} per policy query ($h{=}8$, RTX~4090) with and without \texttt{torch.compile}.}
    \label{tab:latency}
    \begin{tabular}{llcccc}
    \toprule
    & & \multicolumn{2}{c}{\textbf{Latency [s]$\downarrow$}} & \multicolumn{2}{c}{\textbf{Rate [Hz]$\uparrow$}} \\
    \cmidrule(lr){3-4}\cmidrule(lr){5-6}
    \textbf{Model} & \textbf{Size} & default & compiled & default & compiled \\
    \midrule
    VLA-0 (repro.) & 2B   & 3.878 & 1.440 & 2.06 & 5.55 \\
    \midrule
    \multirow{3}{*}{CLAP (Ours)} & 0.8B & 4.233 & \textbf{1.321} & 1.89 & \textbf{6.06} \\
                   & 2B   & 4.307 & \textbf{1.947} & 1.86 & \textbf{4.11} \\
                   & 4B   & 6.013 & \textbf{3.870} & 1.33 & \textbf{2.07} \\
    \bottomrule
    \end{tabular}
  \end{minipage}
\end{table}

\section{Conclusion}
\label{sec:conclusion}
\mai{
We presented CLAP, a minimal VLM-to-VLA fine-tuning recipe that addresses output-distribution mismatch by prepending a natural-language action description to each numeric action sequence. With \diff{4--5 hours of} training on a single 8-GPU node (under 6.5 hours for all model sizes), 2B CLAP achieves 90.8\% on LIBERO, outperforming size-matched VLA-0 by +14.9 points, while also improving OOD robustness on LIBERO-PRO \diff{and outperforming matched VLA-0 on a real UR5e robot}. Multi-scale evaluation across 0.8B, 2B, and 4B reveals that parameter count alone does not determine VLM-to-VLA transfer quality, with the 2B backbone consistently outperforming the 4B variant. 
We believe that making VLA fine-tuning as lightweight and transparent as standard VLM fine-tuning is a necessary step toward bringing the rapid progress of the VLM community to robot learning, and hope that CLAP provides a useful foundation for this direction.
We release all three model scales from a shared Qwen3.5 lineage\diff{, together with code and videos, on the project page}.

}

\section{Limitations}
\label{sec:limitations}
\yuri{\diff{Our study covers two compact VLM families (Qwen3.5 and Qwen3-VL), simulated benchmarks, and a preliminary single-embodiment real-robot study (120 demonstrations); other embodiments and larger real-world datasets remain untested. CLAP does not enforce consistency between the generated prefix and the numeric actions; our ablations suggest success depends on action-grounded semantics rather than prefix precision, but verifying this consistency remains open. CLAP degrades less than VLA-0 rather than fully preserving the pretrained VLM; whether this retains VQA capability is unsettled.} Autoregressive token generation is inherently slower than parallel action decoding (e.g. diffusion or flow-matching heads)\diff{; compiled decoding brings CLAP to 4--6\,Hz at 0.8B--2B, and} LLM deployment techniques such as speculative decoding and quantization are directly applicable but unexplored here.
As CLAP leaves the VLM backbone unchanged, advances such as distillation, quantization, and new architectures can be incorporated without redesigning the action pipeline.}

\acknowledgments{\diff{We thank the anonymous reviewers and the area chair for their constructive feedback, which led to the matched real-robot baseline, the prefix ablations, and the distribution-shift analysis in this version.
This work was supported by the ``Development Acceleration Use'' program of ABCI 3.0, which is provided by AIST and AIST Solutions, and JST PRESTO, Japan, Grant Number JPMJPR2518.}}

\bibliography{draft}  %

\clearpage
\appendix
\section*{Appendix}

\section{Real-Robot Deployment}
\label{app:real-robot}
\diff{This section provides the full setup for the real-robot study summarized in Section~\ref{sec:real_robot}.} The main experiments evaluate CLAP on simulated benchmarks (LIBERO~\citep{liu2023libero}, LIBERO-PRO~\citep{zhou2025liberopro}). 
\yuri{Here we provide a preliminary assessment of whether our direct VLM-to-VLA recipe transfers to a physical robot. We deploy
CLAP on a UR5e manipulator for cluttered pick-and-place, fine-tuning
on a limited real-robot demonstration set to examine how model scale
affects task performance in real-world scenarios.}
\begin{figure}[h]
\centering
\includegraphics[width=0.9\linewidth]{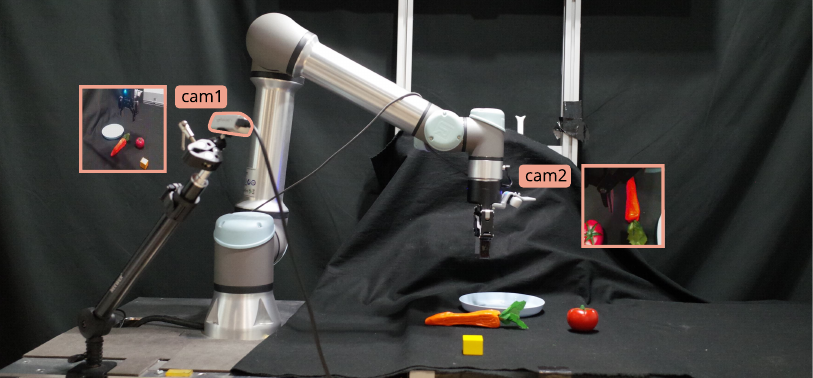}
\caption{\textbf{Real-Robot Setup.} A UR5e arm with a Robotiq 2F-85 gripper performs pick-and-place tasks in cluttered tabletop scenes. Cam1 (fixed side view) and cam2 (wrist-mounted) provide the two RGB observations used as model input. Insets show example frames from each camera.}
\label{fig:real-robot-setup}
\end{figure}
\subsection{Real-Robot Setup}
\Cref{fig:real-robot-setup} shows a real-robot setup to validate our model.
We use a UR5e collaborative robot arm equipped with a Robotiq 2F-85 parallel gripper. A wrist-mounted RealSense D435 provides an eye-in-hand view, and a second RealSense D435 provides a fixed third-person view. CLAP runs on a workstation with an NVIDIA GeForce RTX 4090 GPU, served via FastAPI, and queried by a ROS client at inference time.
The ROS client runs a synchronous observe-query-act loop, at each query, the current side- and wrist-camera images and the language instruction are sent to the inference server via an HTTP POST request, 
which returns an eight-step action chunk.
Each action encodes a Cartesian end-effector delta (position, axis-angle orientation, and gripper state).
The client executes each eight-step action chunk in open loop before re-querying the VLA. At the controller's 15 Hz rate, this yields an effective query rate of $\sim$ 1.9 Hz.
\subsection{Data Collection.}

During data collection, we collect demonstrations on the UR5e via VR controller-based teleoperation. The operator controls the robot using an HTC Vive controller, whose motion is mapped to Cartesian end-effector delta commands at 50 Hz. Data is recorded from two RGB camera streams, consisting of a fixed external camera and a wrist-mounted camera, together with the end-effector pose, gripper state, and a natural-language instruction at each timestep. Each episode lasts up to 1500 steps. For each episode, the scene layout is generated from a seeded episode plan. All three objects (a carrot, a tomato, and a cube) are present simultaneously and are placed at three of five discrete table positions. We use 50 distinct object-to-position assignments from this layout space. Before each demonstration, the operator arranges the scene according to the pre-generated layout.
We collect 120 demonstrations in total, with 40 episodes for each target object. The instruction template is randomized per episode from four templates: \textit{``place \{target\} on the plate,''} \textit{``put \{target\} onto the plate,''} \textit{``pick up \{target\} and place it on the plate,''} and \textit{``grasp \{target\} and drop it onto the plate.''}

\subsection{Training Protocol}
\label{app:real-robot-training}

We fine-tune Qwen3.5-0.8B and 2B directly from their pretrained VLM checkpoints using the CLAP framework on the 120 real-robot demonstrations. \diff{The VLA-0 baseline uses the same Qwen3.5-2B backbone, the same 120 demonstrations, the same one-epoch protocol, and the same inference pipeline.} Notably, we do not initialize from the LIBERO-finetuned models used in the main experiments: because the base VLMs are pretrained on real images, we focus on assessing how much performance can be extracted from VLM-to-VLA adaptation using only real-world data, without an intermediate simulation fine-tuning stage.

\begin{figure}[t]
  \centering
\includegraphics[width=\linewidth]{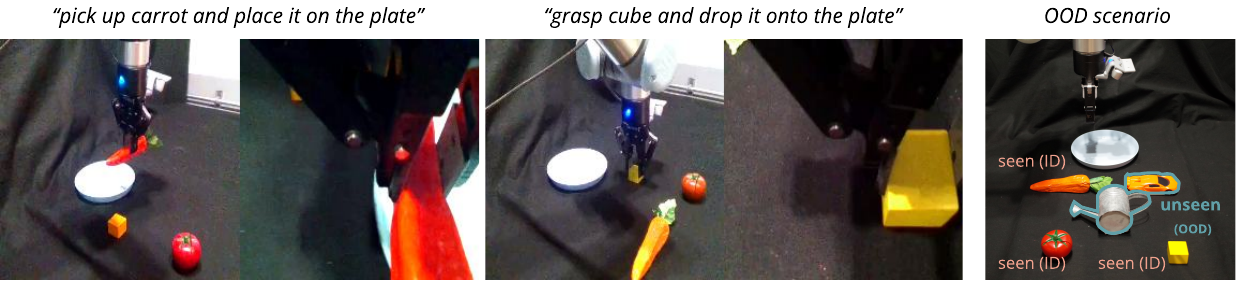}
  \caption{\textbf{Pick-and-Place Task in Cluttered Scene.} Left three: pick-and-place tasks in cluttered scenes with randomized object layouts. Right: an out-of-distribution scenario with unseen objects.}
  \label{fig:real-robot}
\end{figure}

\subsection{Task Design and Evaluation}
\label{app:real-robot-eval}

We evaluate pick-and-place tasks with three training objects (carrot, cube, tomato) under multiple instruction templates. 
\Cref{tab:real-robot-objects} summarizes all objects used in the evaluation.
Each model is tested under two conditions:
\setlist{leftmargin=*}
\begin{itemize}
  \item \textbf{In-domain (ID):} Object sets and layouts seen during training.
  \item \textbf{Out-of-domain (OOD):} Unseen distractor objects (a silver watering can, an orange toy car) are introduced into the scene.
\end{itemize}
\begin{table}[h]
  \small
  \centering
  \caption{Objects used in real-robot evaluation. Seen objects appear in the training demonstrations as pick targets. Unseen objects are introduced as distractors only and are never picked.}
  \label{tab:real-robot-objects}
  \begin{tabular}{llp{7cm}}
    \toprule
    Category & Object & Description \\
    \midrule
    \multirow{3}{*}{Seen (pick target)}
      & \texttt{Carrot}        & Orange toy carrot; elongated shape \\
      & \texttt{Cube}   & Plain yellow wooden cube; uniform color and geometry \\
      & \texttt{Tomato}        & Red toy tomato; spherical shape \\
    \midrule
    \multirow{2}{*}{Unseen (distractor)}
      & \texttt{Silver watering can} & Geometrically complex (spout, handle), dissimilar to any training object \\
      & \texttt{Orange toy car}      & Orange color visually confusable with the carrot, testing robustness to appearance-similar distractors \\
    \bottomrule
  \end{tabular}
\end{table}

Table~\ref{tab:real-robot-results} reports success rates over 20 trials per condition. \diff{The matched VLA-0 baseline reaches 15\% in-domain and 5\% under distractors; we observe that it is less stable under minor environmental variations such as lighting and object placement.} The 2B \diff{CLAP} model achieves the highest success rate (60\%)
followed by 0.8B (35\%),
consistent with simulation results where the 2B backbone also outperforms the 0.8B model.
The overall success rates are substantially lower than in simulation, reflecting the additional complexity of real-world perception and control with only 120 training demonstrations. 
\yuri{Under OOD conditions with unseen distractor objects, the 2B model maintains its in-domain success rate (60\%), while the 0.8B model drops substantially to 10\%, suggesting that the larger backbone is more robust to visual distractors.}
\yuri{This capacity gap is substantially wider than observed on LIBERO (0.8B: 89.6\%, 2B: 90.8\%), suggesting that simulation benchmarks may underestimate the benefit of model capacity, and that real-world task complexity may correlate more strongly with backbone size.}

\begin{table}[h]
  \centering
  \caption{Real-robot pick-and-place success rates (\%). 20 trials per condition.}
  \label{tab:real-robot-results}
  \begin{tabular}{llccc}
    \toprule
    Model & Backbone & Size & ID (\%) & OOD (\%) \\
    \midrule
    \diff{VLA-0 (repro.)} & \diff{Qwen3.5} & \diff{2B} & \diff{15.0} & \diff{5.0} \\
    \midrule
    CLAP & Qwen3.5 & 0.8B & 35.0 & \yuri{10.0} \\
    CLAP & Qwen3.5 & 2B   & 60.0 & \yuri{60.0} \\
    \bottomrule
  \end{tabular}
\end{table}

\mai{The results suggest two complementary directions for future work. The first is to further explore the minimal-recipe regime: improving real-world performance with limited demonstrations through better data augmentation, input robustness, or more effective use of the pretrained VLM's capabilities, without introducing additional architectural components. The second is to scale up robot data while retaining the language-action representation~\cite{raghu2026dontblinkevidencecollapse}, and investigating whether CLAP's zero-modification architecture benefits similarly from larger data remains an open question.}

\section{Inference Latency}
\label{app:inference-latency}

Table~\ref{tab:inference-latency} reports end-to-end inference latency
per policy query, measured as the mean over ten consecutive queries after
one warm-up query on an RTX 4090. The 0.8B and 2B models exhibit
near-identical latency, indicating that inference time is dominated by
token generation overhead rather than model size at these scales. The 4B
model is moderately slower ($\sim$6.0\,s). With $h{=}8$ (open-loop
execution of the full action chunk), effective control rates range from
1.3 to 1.9\,Hz. 

Comparing CLAP 0.8B against VLA-0 0.8B on the same backbone isolates
the overhead of the language-action prefix: CLAP adds approximately
1.0\,s per query (+32\%), corresponding to the additional $\sim$149
tokens generated for the prefix. GPU memory consumption is identical,
confirming that the prefix introduces no architectural
overhead.%
\diff{Table~\ref{tab:latency} in the main text additionally reports latency with \texttt{torch.compile}, which reduces CLAP-0.8B to 1.321\,s (6.06\,Hz) and CLAP-2B to 1.947\,s (4.11\,Hz).}
As CLAP preserves the VLM backbone unmodified, standard LLM
acceleration techniques such as speculative decoding and quantization are
directly applicable and could substantially reduce this overhead.

\begin{table}[h]
  \centering
  \caption{Inference latency per policy query (action horizon $h=8$, RTX 4090). }
  \label{tab:inference-latency}
  \begin{tabular}{llr|cc|c}
    \toprule
    Model & Backbone & Size & Latency (s) & Eff.\ Hz ($k{=}8$) & Peak GPU Memory (GiB) \\
    \midrule
    VLA-0~\cite{goyal2025vla0buildingstateoftheartvlas} & Qwen3.5 & 0.8B & 3.205 & 2.50 & 3.9 \\
    \midrule
    CLAP  & Qwen3.5 & 0.8B & 4.233 & 1.89 & 3.9 \\
    CLAP  & Qwen3.5 & 2B   & 4.307 & 1.86 & 9.1 \\
   CLAP  & Qwen3.5 & 4B   & 6.013 & 1.33 & 18.0 \\   
    \bottomrule
  \end{tabular}
\end{table}

\section{Training Details}
\label{app:hyperparams}
\subsection{Training Hyperparameters}
In Section 5 of the main manuscript, we fine-tune Qwen3.5-0.8,2,4B VLM with full-parameter updates on 8 $\times$ NVIDIA H100 GPUs.
Each GPU processes a micro-batch of 16, yielding a global effective batch size of 128 per optimizer step. \diff{We train for \textbf{17k gradient steps} (about 2.18M training samples) on the LIBERO-\{spatial, object, goal, long\} training set under our augmentation pipeline. The original VLA-0 code does not use a distributed sampler, so with 8 GPUs each GPU iterates over the full dataset and 17k steps amount to 1 epoch there; our released implementation uses a distributed sampler, under which the same 17k steps correspond to 8 epochs. We therefore specify the budget in gradient steps throughout.} 
\begin{table}[h]
    \centering
    \footnotesize
    \caption{Training hyperparameters shared across all CLAP and VLA-0 experiments.}
    \begin{tabular}{ll}
    \toprule
    \textbf{Hyperparameter} & \textbf{Value} \\
    \midrule
    Optimizer               & AdamW \\
    Training steps          & 17k\\
    Learning rate           & $5.0\times10^{-6}$ (constant schedule) \\
    Weight decay            & $10^{-10}$ \\
    Effective batch size    & 128 (16 per GPU $\times$ 8 GPUs, DDP) \\
    Mixed precision         & bfloat16 \\
    Attention               & Flash Attention~2 \\
    Image augmentation      & Random crop (scale 0.875), color jitter \\
    Action masking rate     & 0.4\\
    \bottomrule
    \end{tabular}
\end{table}

\subsection{Loss Convergence vs. Task Saturation}
\begin{figure}[h]
  \centering
  \diff{\includegraphics[width=\linewidth]{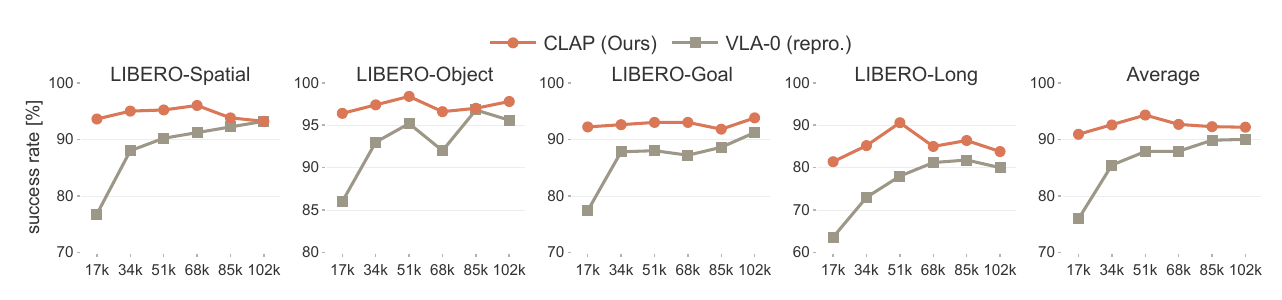}
  \caption{\textbf{LIBERO success rate versus training steps} for CLAP and VLA-0 on Qwen3.5-2B; the x-axis is gradient steps. The 17k-step points come from an independent evaluation run and differ from Table~\ref{tab:libero} by at most 1.4\,pt per suite. CLAP reaches about 90\% average success at 17k steps, the level VLA-0 reaches only at 102k steps.}
  \label{fig:libero_steps}}
\end{figure}

Section 5.2 of the main text reports LIBERO performance after \diff{17k gradient steps}
(\diff{4--5 hours} on 8$\times$H200 GPUs). 
\diff{Figure~\ref{fig:libero_steps} extends training of both CLAP and VLA-0 (Qwen3.5-2B) up to 102k gradient steps (6$\times$17k).
CLAP already reaches 90.9\% average success at 17k steps, a level VLA-0 attains only at 102k steps (90.0\%), i.e., with six times the training budget.
CLAP peaks at 94.3\% at 51k steps and then plateaus around 92\%, whereas VLA-0 keeps improving throughout.
We therefore adopt the 17k-step checkpoint throughout, which isolates the learning-efficiency gap between the two output representations.}

\section{Evaluation Protocol}
Each observation encodes two camera views (agentview and wrist) tiled into a single $224\!\times\!224$ image.
The model produces the full \texttt{<think>}\ldots\texttt{</think>}\,\texttt{action\_tokens} sequence in one autoregressive pass using Qwen3.5's built-in extended-thinking mode (\texttt{enable\_thinking=True}); only the 56 integer tokens after the closing \texttt{</think>} tag are forwarded to the robot controller.
All eight predicted action steps execute open-loop before the model is re-queried (action horizon $h{=}8$).

\section{CLAP Language Target Construction}
\label{app:template}
\diff{Table~\ref{tab:action-example} shows a complete example of the system message, instruction, and generated output sequence.}
\begin{table}[h]
\footnotesize
\begin{tabularx}{\linewidth}{X}
\toprule[1.1pt]
\textbf{System Message:}\\ %
\ttfamily
Analyze the input image and predict robot actions for the next 8
timesteps. Each action has 7 dimensions. First describe the motion in
natural language inside \colortag{teal}{think}...\colortag{teal}{/think}. Then output exactly 56
integers (0-1000 each), space-separated, for the discretized actions in
timestep order. Nothing else.\\
\midrule
\textbf{User Instruction:} 
\ttfamily grasp cube and drop it onto the plate. \\
\midrule[1pt]
\textbf{Language-Action Prediction:}\\
\ttfamily
\colortag{teal}{think}
move back 0.5 cm, move up 3.5 cm, move left 0.4 cm, tilt left 1 degrees, tilt back 1 degrees, rotate counterclockwise 3 degrees, open gripper
\colortag{teal}{/think}\\
\footnotesize\ttfamily 408 980 166 319 505 491 1000 407 980 159 312 496 486 1000 406 980 152 303 486 479 1000 406 981 150 300 489 480 1000 407 980 149 294 484 483 1000 408 980 146 289 482 480 1000 410 979 144 287 486 483 1000 412 978 137 277 484 479 1000
\\
\bottomrule[1.1pt]
\caption{
\textbf{Example CLAP output sequence.}
Given an instruction, system message, and image observations, CLAP first generates a fixed-template language-action prefix that summarizes the intended $h$-step motion, then generates discrete numeric action tokens for execution.
The prefix provides a structured intermediate conditioned on the observation and instruction, while the numeric tokens encode the precise 7-DoF command for each step.
}\label{tab:action-example}
\end{tabularx}
\end{table}

During training, the language-action prefix is generated
deterministically from each ground-truth action chunk using a fixed
template-based method~\cite{zha2026laplanguageactionpretrainingenables} that summarizes steps of numerical actions into a short natural-language
summary.

\paragraph{Inputs.}
Let $\bm a \in \mathbb R^{7 \times h}$ denote an action
chunk of horizon $h$, where each $a_t = (\delta x_t,
\delta y_t, \delta z_t, \delta \phi_t, \delta \theta_t, \delta
\psi_t, g_t)$ contains end-effector translational deltas in meters,
rotational deltas in radians (roll, pitch, yaw), and a gripper
state $g_t \in \{0, 1\}$.

\paragraph{Aggregation.}
We sum the per-step deltas across the chunk and convert to
human-readable units:
\begin{align}
\Delta x &= \mathrm{round}\!\left(100 \sum_{t=1}^{h} \delta x_t\right)
  \,\,\text{[cm]},
  \qquad &\text{and analogously for $\Delta y, \Delta z$,} \\
\Delta \phi &= \mathrm{round}_{10}\!\left(\frac{180}{\pi}
  \sum_{t=1}^{h} \delta \phi_t\right) \,\,\text{[deg]},
  \qquad &\text{and analogously for $\Delta \theta, \Delta \psi$,}
\end{align}
where $\mathrm{round}_{10}$ rounds to the nearest multiple of $10^{\circ}$.
The gripper bit is taken from the last step,
$g_H \geq 0.5 \Rightarrow \texttt{open}$, otherwise \texttt{close}.

\paragraph{Phrasing.}
Each non-zero aggregated component is mapped to a phrase whose
direction word is selected by the sign of the underlying delta:
\begin{center}
\small
\begin{tabular}{lll}
\toprule
Axis & Sign $> 0$ & Sign $< 0$ \\
\midrule
$\Delta x$ & \texttt{move forward $|\Delta x|$ cm} &
             \texttt{move back $|\Delta x|$ cm} \\
$\Delta y$ & \texttt{move left $|\Delta y|$ cm} &
             \texttt{move right $|\Delta y|$ cm} \\
$\Delta z$ & \texttt{move up $|\Delta z|$ cm} &
             \texttt{move down $|\Delta z|$ cm} \\
$\Delta \phi$ & \texttt{tilt left $|\Delta\phi|$ degrees} &
                \texttt{tilt right $|\Delta\phi|$ degrees} \\
$\Delta \theta$ & \texttt{tilt back $|\Delta\theta|$ degrees} &
                  \texttt{tilt forward $|\Delta\theta|$ degrees} \\
$\Delta \psi$ & \texttt{rotate counterclockwise $|\Delta\psi|$ degrees} &
                \texttt{rotate clockwise $|\Delta\psi|$ degrees} \\
\bottomrule
\end{tabular}
\end{center}
The selected phrases, followed by the gripper phrase, are joined with
\texttt{", "} into a single sentence $d$.

\paragraph{Final target.}
The CLAP target is the language summary wrapped in
\verb|<think>...</think>| tags and followed by the same discretised
numeric action tokens used by VLA-0:
\[
\texttt{<think>}\,d\,\texttt{</think>}\,\,
n_1\,n_2\,\ldots\,n_{7\cdot h}
\]
where $n_i$ are the per-dimension action bin indices. Concretely,
for $h=8$ with a small forward and downward motion and the gripper
closing, the target looks like
\begin{center}
\small
\verb|<think>move forward 3 cm, move down 2 cm, close gripper</think> 17 13 ... 0|
\end{center}
At inference, the prefix up to and including the last
\verb|</think>| is stripped and the remaining numeric tokens are
decoded back to a continuous action chunk.

\section{Action Masking Details}
\label{app:masking}
\diff{This table was moved here from the main text; masking is an optional augmentation and not part of the core CLAP recipe.}
\begin{table}[h]
    \centering
    \scriptsize
    \caption{\textbf{Action masking ablation} on LIBERO (\diff{17k steps}, $h{=}8$), discussed in Section~\ref{sec:learning_efficiency}.
             We highlight the \hcb{best} per size group.
             \ding{51} indicates action masking; -- indicates no masking. Parentheses in Avg show the change from unmasked CLAP.}
    \label{tab:ablation}
    \begin{tabular}{lllcccccl}
    \toprule
    \textbf{Model} & \textbf{Backbone} & \textbf{Size} & \textbf{Masking}
      & \textbf{Spatial$\uparrow$}
      & \textbf{Object$\uparrow$}
      & \textbf{Goal$\uparrow$}
      & \textbf{Long$\uparrow$}
      & \textbf{Avg$\uparrow$} \\
    \midrule
    CLAP &Qwen3.5 & 0.8B & --         & \hcb{\textbf{92.6}} & \hcb{\textbf{97.8}} & 88.4 & \hcb{\textbf{79.6}} & \hcb{\textbf{89.6}} \\
    &        &      & \ding{51}  & 87.6 & 95.4 & \hcb{\textbf{90.0}} & 69.8 & 85.7 {\color{red}($-3.9$)} \\
    \midrule
    CLAP& Qwen3.5 & 2B   & --         & 93.0 & \hcb{\textbf{97.4}} & 90.8 & \hcb{\textbf{82.0}} & \hcb{\textbf{90.8}} \\
    &        &      & \ding{51}  & \hcb{\textbf{94.8}} & 96.8 & \hcb{\textbf{91.6}} & 73.2 & 89.1 {\color{red}($-1.7$)} \\
    \midrule
    CLAP &Qwen3.5 & 4B   & --         & 88.0 & \hcb{\textbf{97.4}} & 86.6 & 67.6 & 84.9 \\
     &       &      & \ding{51}  & \hcb{\textbf{91.6}} & 92.2 & \hcb{\textbf{89.8}} & \hcb{\textbf{78.8}} & \hcb{\textbf{88.1}} {\color{green}\textbf{(+3.2)}} \\
    \bottomrule
    \end{tabular}
\end{table}

Following prior work~\cite{goyal2025vla0buildingstateoftheartvlas}, each
action chunk is discretized into $1000$ bins per dimension and
serialized as space-separated integers. Cross-entropy loss is computed
only over the assistant turn; all prompt-side tokens (system message,
images, instruction, and padding) are masked out. As an optional
augmentation, a random fraction of input action tokens are replaced with
a \texttt{?} placeholder while keeping the original values as prediction
targets, forcing the model to reconstruct masked actions from visual and
linguistic context. 
For each sample, the masking fraction is drawn from
$\mathrm{Uniform}(0,\, p_{\max})$ with $p_{\max}{=}0.4$. 
For CLAP, this augmentation
is restricted to the numeric tail following \colortag{teal}{/think} to avoid
misalignment within the language-action prefix. At inference, a logits
processor constrains generation to digits, spaces, and \texttt{eos}; in
CLAP mode this constraint is suspended during the \colortag{teal}{think} span
and re-enabled for the numeric tail.

\section{Benchmark Details}
\label{app:benchmarks}

\paragraph{LIBERO.}
LIBERO~\citep{liu2023libero} is a simulated manipulation benchmark
comprising four task suites: Spatial, Object, Goal, and Long.
Each suite contains 10 tasks with 50 demonstrations per task.
We follow the standard evaluation protocol and report success rates
averaged over 20 rollouts per task.

\paragraph{LIBERO-PRO.}
LIBERO-PRO~\citep{zhou2025liberopro} extends LIBERO with four perturbation types designed to stress-test out-of-distribution robustness.
\textit{Obj} replaces objects with novel visual instances of the same category.
\textit{Pos} relocates objects to positions unseen during training.
\textit{Sem} rephrases task instructions while preserving semantics.
\textit{Task} modifies the task configuration itself (e.g., goal location or object arrangement), representing the hardest perturbation type. All four perturbation types are applied to each of the four LIBERO suites.

\paragraph{VLABench.}
\yuri{VLABench~\citep{liu2024vlabench} evaluates task planning across
45 tasks covering six capability dimensions:
Common Sense \& World Knowledge (background knowledge about object properties and everyday scenarios),
Spatial Understanding (precise spatial reasoning about object positions and orientations),
Mesh \& Texture Understanding (distinguishing objects by shape or surface appearance),
Semantic Understanding (language-grounded category-level reasoning),
Physics Law (tasks governed by physics constraints such as gravity, support, and rigidity), and
Complex (multi-step inference combining multiple dimensions).
Each dimension covers 100 episodes; results are averaged over all episodes within each dimension.}

\begin{figure}[t]
  \centering
  \includegraphics[width=\linewidth]{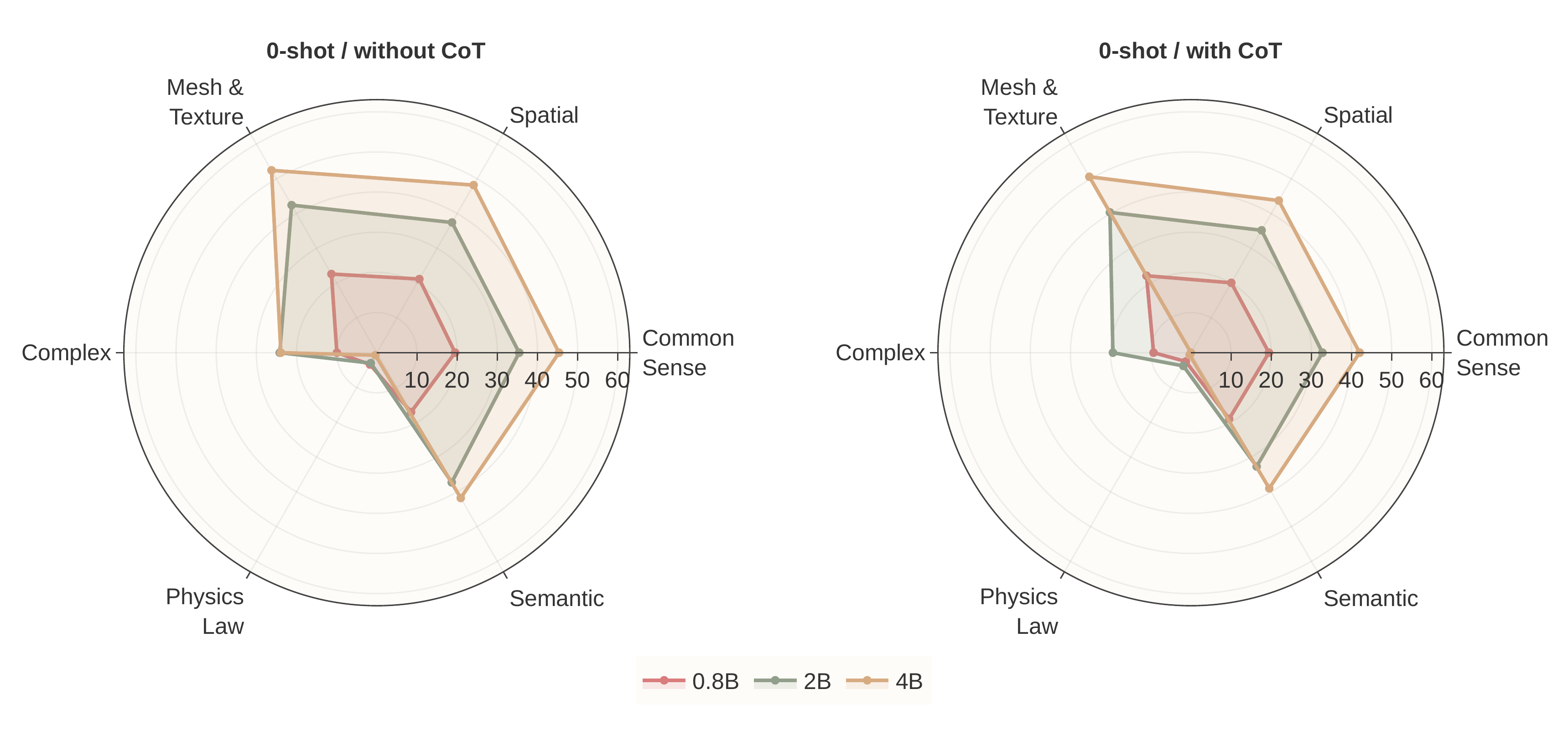}
  \caption{
    \yuri{\textbf{VLABench capability profiles under 0-shot prompting.}
    Same evaluation as Figure~3 in the main paper (1-shot) but without an in-context example.
    The overall capability ranking across model scales is consistent with the 1-shot setting, with the 4B model showing balanced performance across most dimensions.
    However, Physical Law accuracy, which remains modest but nonzero under 1-shot prompting (Figure~3 in the main paper), drops to near zero for all three scales without the in-context example, suggesting that even this limited performance relies on the provided demonstration.
    }
    }
  \label{fig:vlabench_0shot}
\end{figure}

\yuri{The main paper reports 1-shot evaluation of the Qwen3.5 backbones before robot fine-tuning. We repeat the evaluation under 0-shot prompting (Figure~\ref{fig:vlabench_0shot}) and find that the overall capability ranking is consistent, although Physical Law accuracy drops to near zero without the in-context example.}

\section{SmolVLA Full Size Comparison}
\label{app:smolvla}

\yuri{Table~\ref{tab:smolvla_full} reports all three SmolVLA sizes
alongside CLAP for reference.
SmolVLA uses community-scale robot pretraining, whereas CLAP
uses no robot pretraining beyond the LIBERO demonstrations.
Despite this difference, CLAP 0.8B and 2B match or exceed all
three SmolVLA scales on average success rate with \diff{17k-step}
training, while CLAP 4B falls below SmolVLA 2.25B, consistent
with the non-monotonic scaling observed in the main paper.}

\begin{table}[h]
  \centering
  \scriptsize
  \caption{SmolVLA all sizes vs.\ CLAP on LIBERO (\diff{17k steps},
           $h{=}8$).
           {\color{refgray}Gray} rows show SmolVLA under its
           full training protocol (contextual reference only).}
  \label{tab:smolvla_full}
  \begin{tabular*}{\textwidth}{@{\extracolsep{\fill}}llll cccc l}
  \toprule
  \textbf{Model} & \textbf{Backbone} & \textbf{Size}
    & \textbf{VLA Pretrain}
    & \textbf{Spatial$\uparrow$}
    & \textbf{Object$\uparrow$}
    & \textbf{Goal$\uparrow$}
    & \textbf{Long$\uparrow$}
    & \textbf{Avg$\uparrow$} \\
  \midrule
  \rowcolor{magenta!20}\multicolumn{9}{l}{\textit{Ours (\diff{17k steps}, identical protocol)}} \\
  \quad CLAP & Qwen3.5 & 0.8B & --
    & 92.6 & 97.8 & 88.4 & 79.6 & 89.6 \\
  \quad CLAP & Qwen3.5 & 2B & --
    & 93.0 & 97.4 & 90.8 & 82.0 & 90.8 \\
  \quad CLAP & Qwen3.5 & 4B & --
    & 88.0 & 97.4 & 86.6 & 67.6 & 84.9 \\
  \midrule\midrule
  \rowcolor{yellow!10}\multicolumn{9}{l}{\rv{\textit{SmolVLA (full training protocol)}}} 
  \\
  \quad \rv{SmolVLA$^\dagger$} & \rv{SmolVLM2} & \rv{0.24B} & --
    & \rv{87.0} & \rv{93.0} & \rv{88.0} & \rv{63.0} & \rv{82.8} \\
  \quad \rv{SmolVLA$^\dagger$} & \rv{SmolVLM2} & \rv{0.45B} & --
    & \rv{90.0} & \rv{96.0} & \rv{92.0} & \rv{71.0} & \rv{87.3} \\
  \quad \rv{SmolVLA$^\dagger$} & \rv{SmolVLM2} & \rv{2.25B} & --
    & \rv{93.0} & \rv{94.0} & \rv{91.0} & \rv{77.0} & \rv{88.8} \\
  \bottomrule
  \end{tabular*}
  \vspace{2pt}
  \scriptsize
  $\dagger$ SmolVLA uses community-scale robot pretraining.
\end{table}

\newpage

\end{document}